%% file: iclr2025_conference.tex
\documentclass{article} 
\usepackage{iclr2025_conference,times}

\input{math_commands.tex}

\usepackage{hyperref}
\usepackage{url}

\usepackage{graphicx}
\usepackage{subfigure}
\usepackage{booktabs} 

\usepackage{amsmath}
\usepackage{amssymb}
\usepackage{mathtools}
\usepackage{amsthm}
\usepackage{float}
\usepackage{wrapfig}

\usepackage{mdframed}

\definecolor{mygray}{gray}{0.98}
\definecolor{mycolor}{rgb}{1, 0.992, 0.961}

\newmdenv[
backgroundcolor=mygray, 
linecolor=black, 
linewidth=1pt, 
skipabove=1em, 
skipbelow=1em, 
innerleftmargin=10pt, 
innerrightmargin=10pt, 
innertopmargin=10pt, 
innerbottommargin=10pt
]{beautifulbox}

\newcommand{\myfont}{lmr}

\title{Empirical Asset Pricing with Large Language Model Agents}

\author{Junyan Cheng \\
Thayer School of Engineering\\
Dartmouth College\\
Hanover, NH 03755, USA \\
\texttt{jc.th@dartmouth.edu} \\
\And
Peter Chin \\
Thayer School of Engineering\\
Dartmouth College\\
Hanover, NH 03755, USA \\
\texttt{pc@dartmouth.edu} 
}

\iclrfinalcopy 
\begin{document}

\maketitle

\begin{abstract}
In this study, we introduce a novel asset pricing model leveraging the Large Language Model (LLM) agents, which integrates qualitative discretionary investment evaluations from LLM agents with quantitative financial economic factors manually curated, aiming to explain the excess asset returns. The experimental results demonstrate that our methodology surpasses traditional machine learning-based baselines in both portfolio optimization and asset pricing errors. Notably, the Sharpe ratio for portfolio optimization and the mean magnitude of $|\alpha|$ for anomaly portfolios experienced substantial enhancements of 10.6\% and 10.0\% respectively. Moreover, we performed comprehensive ablation studies on our model and conducted a thorough analysis of the method to extract further insights into the proposed approach. Our results show effective evidence of the feasibility of applying LLMs in empirical asset pricing.
\end{abstract}

\section{Introduction}

The pricing of financial instruments, including equities, has occupied a central position in empirical financial economics studies. This research exerts a notable influence on societal welfare by promoting Pareto efficiency in the distribution of capital. Present asset valuation techniques typically involve the meticulous development of manual macroeconomic indicators or factors specific to individual companies to forecast future excess returns \cite{FF3,FF5}. Although these methods have achieved considerable success in practical markets, they face criticism from the Efficient Market Hypothesis (EMH). The EMH suggests that in an efficient market, the predictive power of manually designed factors diminishes as market participants eventually uncover and utilize these predictors.

Due to this rationale, linguistic data, which are the primary sources of traditional discretionary investing, become essential. This is attributed to the substantial influence of linguistic information flow on societal and market dynamics. The importance of this is further underlined in actual financial settings, where discretionary portfolio management continues to play a crucial role \cite{MVM}. Investment decisions in this realm are predominantly informed by the manager's expertise and intuition, with assets being assessed and appraised on the basis of data derived from news, inquiries, and reports, among other sources, rather than relying on quantitative modeling.

This phenomenon emphasizes two significant aspects. First, qualitative discretionary analyses have the potential to reveal crucial pricing information that is not present in traditional economic indicators or market datasets. Second, despite the incorporation of current natural language processing and semantic analysis approaches, quantitative factor models have yet to fully integrate these insights. Realizing the synergy between these approaches remains a challenging yet enticing goal \cite{cao2021man}. Nevertheless, the use of linguistic information is compounded by the need for financial reasoning and the ability to maintain a long-term memory of tracking events and company perceptions for accurate interpretation. Moreover, suboptimal interactions in model architecture between linguistic and manual elements may result in noise instead of information gain \cite{NAP}.

\begin{figure*}[!htb]
    \centering
    \includegraphics[width=\linewidth]{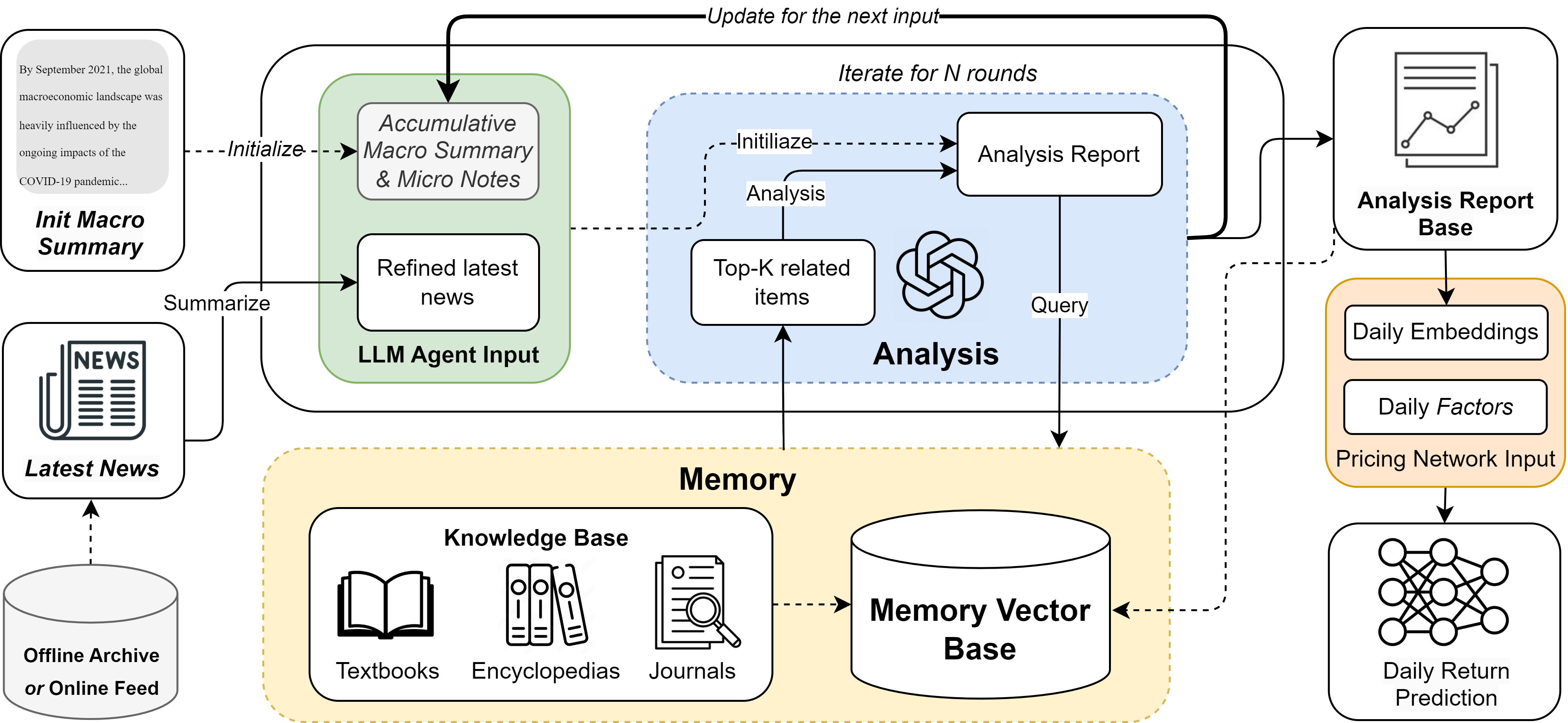}
    \caption{
    The LLM agent produces analysis reports from the latest news through a multi-step refinement, incorporating past reports and domain knowledge from memory. For simplicity, the filter for irrelevant news is excluded. A macro and micro note, continuously updated by the latest analysis report, is used to provide additional context. The average embedding of daily analysis reports will be input into the pricing network along with daily manual factors.
    }
    
    \label{fig:main}
\end{figure*}

Recent advancements in Large Language Models (LLMs) promote the utilization of textual and other alternative data in financial domains \cite{sociodojo}. However, most existing methods focus on the trading strategies for individual stocks \cite{finagent,finvision}, other than the macroscopic asset pricing problem which aims to discover the explanatory factors across all assets in the market rather than optimizing for the prediction on a single specific company. 

In this study, we introduce a novel LLM agent asset pricing model, which fuses discretionary investment analysis simulated by an LLM agent and quantitative factor-based methods. Our method employs the LLM agent to iteratively analyze the latest news, supported by a memory of previous analysis reports and a knowledge base comprising books, encyclopedias, and journals. The embedding of analysis reports is merged with manual factors to predict future excess asset returns. Besides offering a performance edge, our method also provides enhanced interpretability through generated analysis reports.
We evaluate our approach with a dataset consisting of three years of news and approximately 70 years of economic and market data. The experimental results show that our approach surpasses machine learning-based asset pricing baselines, achieving a 10.6\% increase in the Sharpe ratio for portfolio optimization and a 10.0\% improvement in the average $|\alpha|$ for asset pricing errors in character-section portfolios.
Our work demonstrates the viability of applying LLMs for asset pricing.
The primary contributions are summarized as follows:
\begin{itemize}
\item Introduced a novel LLM agent architecture to analyze business news for discretionary investment insights as pricing signals.
\item Proposed a hybrid asset pricing framework that combines qualitative discretionary analysis and quantitative manual factors.
\item Performed comprehensive experiments to assess the effectiveness of the proposed approach with in-depth analysis of components. 
\end{itemize}
Our code and data can be found in \url{https://github.com/chengjunyan1/AAPM}.

\section{Related Work}

\paragraph{Asset Pricing for Security.}
Asset pricing aims to search for the fair price of financial assets, such as securities, by discovering underlying explanatory factors across different assets. 
\citet{CAPM} introduced the groundbreaking Capital Asset Pricing Model (CAPM), which breaks down the expected return of an asset into a linear function of the market return. Various extensions of the CAPM have been developed. \citet{ICAPM} incorporated wealth as a state variable, while \citet{CCAPM} considered consumption risk as a pricing factor. The single-factor CAPM was later expanded into multi-factor models.  \citet{FF3} proposed the Fama-French 3-factor (FF3) model, which explains returns by size, leverage, book-to-market equity, and earnings-price ratios. They later revised it to a 5-factor model \cite{FF5}. Furthermore, \citet{FF4} identified momentum as an additional factor. \citet{APT} formulated the Arbitrage Pricing Theory (APT), which considers asset pricing as an equilibrium in the absence of arbitrage opportunities. The Stochastic Discount Factor (SDF) calculates the price by discounting future cash flows using a stochastic pricing kernel \cite{SDF}. We introduce LLMs to produce news-based macroeconomic factors for additional explanations of prices.

\paragraph{Financial Machine Learning.}
The application of machine learning techniques has been introduced to explore the non-linear interactions among the growing ``factor zoo'' \cite{FactorZoo}. Instrumented Principal Component Analysis (IPCA) was developed by \citet{IPCA} to estimate latent factors and their loadings from data. \citet{NNAP} introduced a deep neural network to model interactions. \citet{CA} proposed a conditional autoencoder that considers latent factors and asset characteristics as covariates. \citet{DLAP} utilized Generative Adversarial Networks to train a neural SDF based on the methods of moments. 
\cite{NBERw33351} introduced a linear transformer which further reduces the pricing errors.
Furthermore, \citet{bybee2021business} conducted an analysis of the Wall Street Journals (WSJ) to gauge the state of the economy. 
Based on this analysis, \citet{NAP} further suggested using Latent Dirichlet Allocation (LDA) to analyze monthly news topics from WSJ as pricing factors. Recent NLP methods \cite{xu2018stock,xie2022word} have been employed to forecast stock movements, in contrast to asset pricing, they do not aim to find interpretable factors that explain anomalies in excess asset returns. Our LLM-based approach offers an alternative interpretation through analysis reports.

\begin{wrapfigure}{r}{0.5\textwidth}
    \begin{center}
        \centering
                \includegraphics[width=0.48\linewidth]{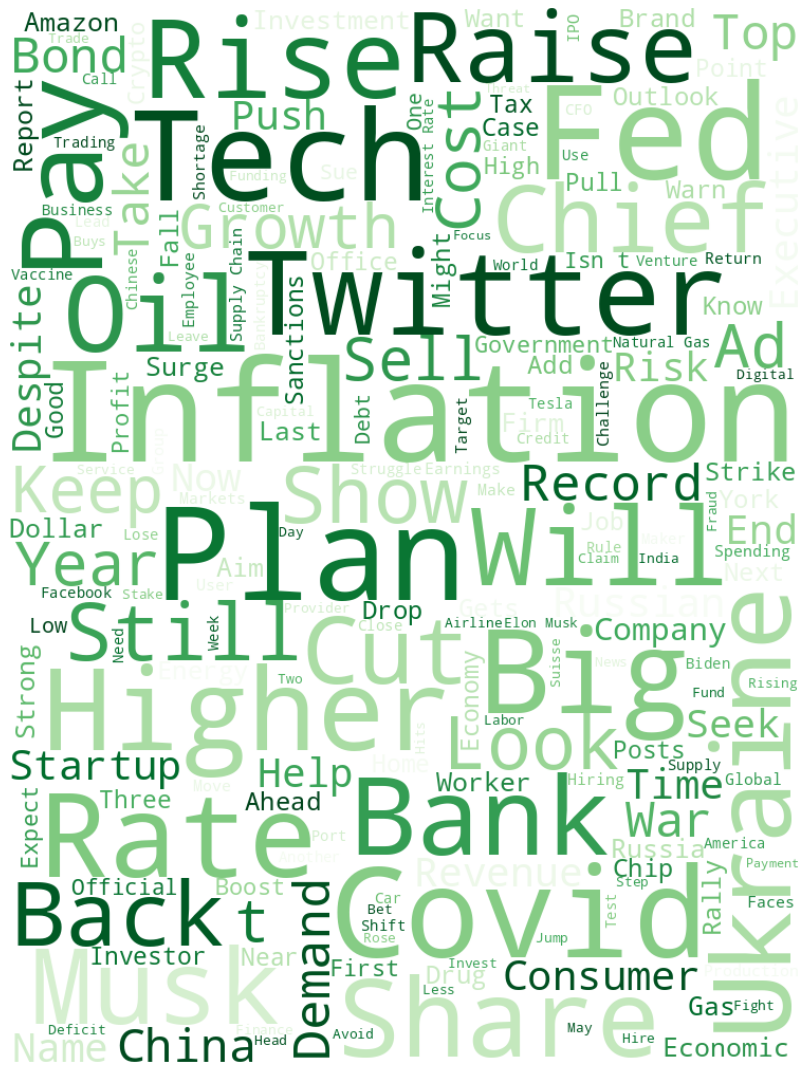}
                \includegraphics[width=0.48\linewidth]{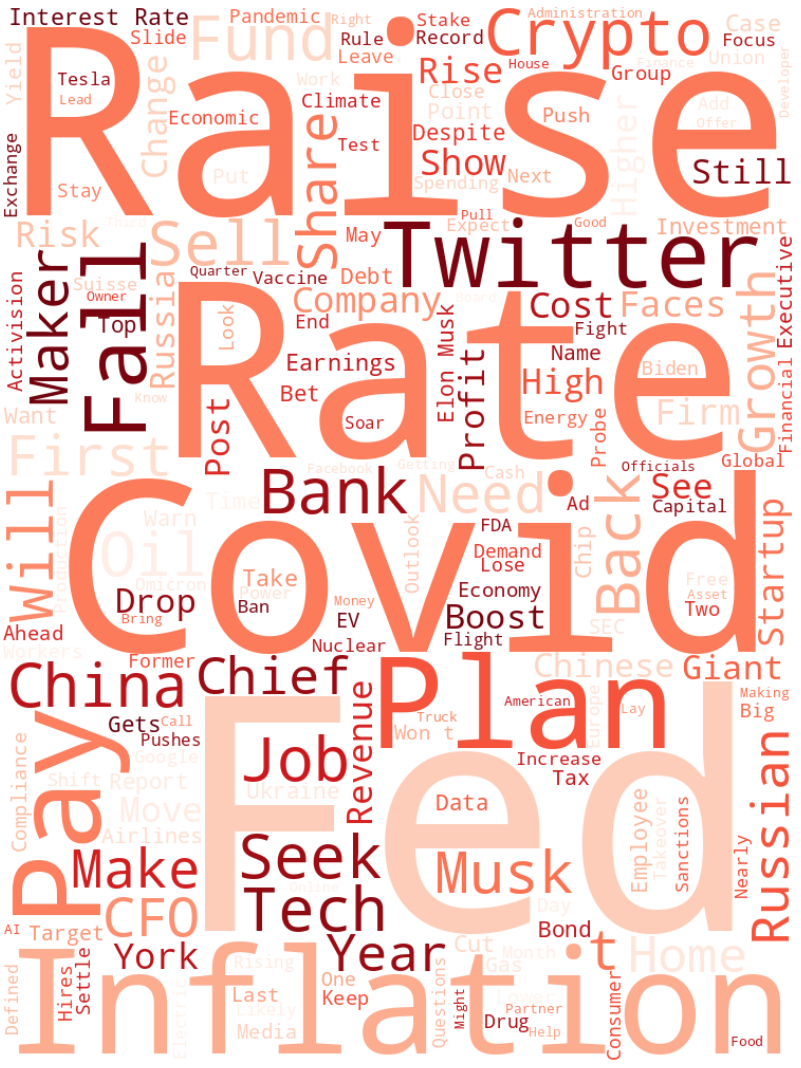}
                \caption{Visualization of the keywords in the titles of news articles on the days over the highly complicated first two-year span in our dataset when the market return is positive (left) and negative (right).}
            \label{fig:pos_neg_kw}
    \end{center}
\end{wrapfigure}

\paragraph{Large Language Model Agents for Finance.}
LLM agents possess powerful emergent capabilities, such as reasoning, planning, and tool-using \cite{GPT4}. The core of LLM agent programming lies in prompting, which employs contextual hinting text to regulate the output of LLM \cite{liu2023pre}. Several prompting strategies have been proposed. Chain-of-Thoughts (CoT) \cite{COT} encourages the agent to reason in a step-by-step manner. \citet{yao2022react} introduced the ReAct prompt, enabling the agent to refine its output based on the results of previous attempts. It allows the agent to use external tools, such as databases and search engines. Memory is another crucial component of LLM agents. \citet{hu2023chatdb} introduced databases as symbolic memories. \citet{packer2023memgpt} stores dialogues in both long- and short-term memory, analogously to operating systems. \citet{sociodojo} developed an agent capable of making ``investment'' decisions on social science time series based on input news, reports, etc., and knowledge base, as well as the Internet. We focus on using the agent to simulate discretionary investment decision-making to synergize qualitative and quantitative asset pricing.
LLM Agent has been applied for stock prediction, \citet{koa2024learning,ding2024tradexpert,finagent,finvision} proposed multi-LLM agents to predict the movement direction of specific stocks while we focus on the fair pricing of all stocks in the market.

\section{Method}
\label{sec:method}

Given a state vector $V_{\tau,a}$ at a time point $\tau\in \{0,1,2,...\}$, which represents the current status of the market, society, and an asset $a$, an asset pricing model predicts the excess returns $r_{\tau+1,a}$ of the asset at the subsequent time point, expressed as $P(r_{\tau+1,a}|V_{\tau,a})$.  In our study, each time point corresponds to one day. 
In traditional factor-based methods, the state $V_{\tau,a}\in\mathcal{N_F}^N$ is a vector composed of $N_F$ factors that are manually derived from economic indicators, market data, asset characteristics, etc. For instance, the market excess return, the performance disparity between small and large firms, and the difference between high and low book-to-market companies in the Fama-French 3-factor model \cite{FF3}. Recently, \citet{bybee2021business} demonstrated that a collection of business news can serve as an alternative representation of macroeconomic conditions, while \citet{NAP} employs LDA to extract news characteristics as economic predictors for pricing. Building on this idea, we use the average embedding of analysis reports that mine values from the news as a proxy for the society, economic, and market states.

Business news in major media outlets like the WSJ carries important market information, however, they typically restrict their interpretations and opinions, leaving room for discretionary analysis. It is crucial to understand that business events are often interrelated.

As visualized in Figure \ref{fig:pos_neg_kw} about the keywords found in the titles of the news articles on days with positive and negative market returns. It corresponds well with human intuition about how the market trend was driven, long-term events like the FED rate hike, COVID, and inflation worries have had the most significant negative effects on the market over the two-year span from Sep. 2021 to Sep. 2023 of our dataset, whereas elements such as technology, Twitter, and inflation control measures have driven market growth.  Interpreting business news about such key events requires an extrapolation process that depends on extensive background knowledge and historical events. 

Based on these observations, we introduce an LLM agent asset pricing model, utilizing an LLM agent with long-term memory of domain knowledge and historical news analysis to iteratively analyze the input news and generate the analysis report, as detailed in Section \ref{ssec:main}. Subsequently, we combine these qualitative analysis reports and quantitative manual factors to feed into our hybrid asset pricing network in Section \ref{ssec:hapm}.

\begin{wrapfigure}{r}{0.5\textwidth}
    \begin{center}
            \centering
            \includegraphics[width=\linewidth]{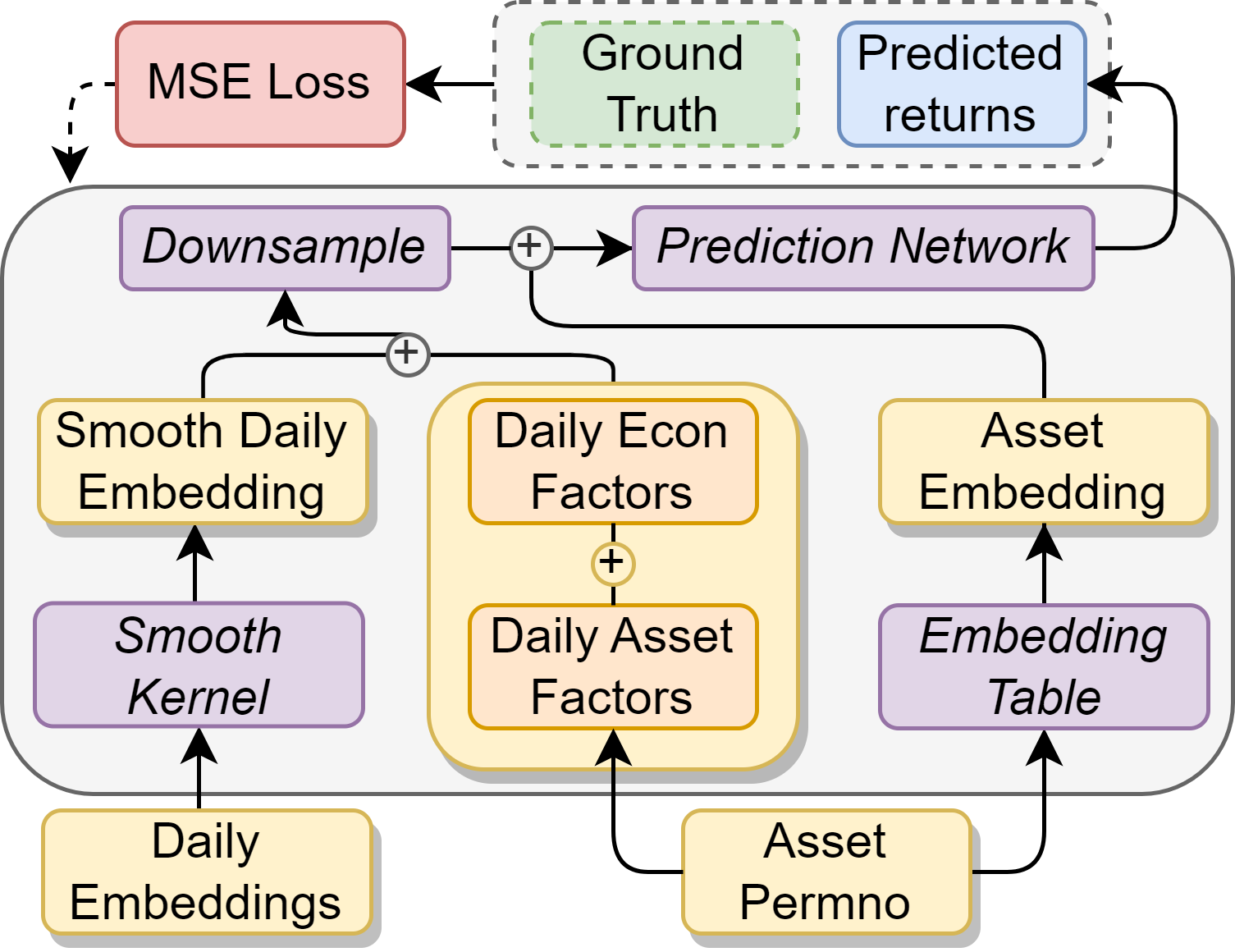}
            \caption{The demonstration of our hybrid asset pricing network. The purple boxes mark the computational components. Yellow boxes are data, and the circled plus symbol means concatenation. The MSE loss is computed with predicted returns feedback to update the network.}
            \label{fig:hap}
    \end{center}
\end{wrapfigure}

\subsection{Discretionary Analysis with LLM Agent}
\label{ssec:main}

The agent utilizes the latest news $x_{t}$ at time $t$ (e.g., a WSJ article published at 9:32 AM on 6 June 2020), along with a note $n_{t}$ on macroeconomics and market trends, to generate an analysis report $R_{t}$. The note $n_{t}$ is initialized with a macroeconomics summary $n_0$ produced by an LLM, before a cut-off date $d_k$.
It offers necessary macroscopic context on economic and societal trends not directly available from the news or the memory. The note is then iteratively updated to $n_{t'}$ with the new analysis report $R_{t}$ to keep the context up-to-date, and we also prompt the agent to document investment ideas and market thoughts in the notes to provide a short-term background such as the trends on the market, long-term research opportunities to watch. To ensure the note is continuously updated without missing information, the dataset in our study starts from $d_k+1$, immediately following the cut-off date.

The analysis process begins with generating a refined news item $x'_{t}$ that summarizes key information from the raw input $x_t$. This step helps control the input length and standardizes the format and style. The refined news $x'_{t}$ and the note $n_{t}$ are then combined to form an input $I_{t}$ for the agent. The agent will determine if the news contains investment information: if not, it will be skipped; otherwise, an initial analysis report $R_{t}^{0}$ will be created. The report undergoes iterative refinement over $N$ rounds. In each round $i$, the report $R_{t}^{i-1}$ is used to query an external memory $M^t$, a vector database initialized with the SocioDojo knowledge base \cite{sociodojo}, which includes textbooks, encyclopedias, and academic journals in fields such as economics, finance, business, politics, and sociology. We use BGE \cite{bge_embedding} as the embedding model $f_e$, which maps text to a vector $e\in\mathcal{R}^{d_{emb}}$ for querying the memory. This choice is based on the MTEB leaderboard \cite{MTEB}, where we selected the best retrieval model considering performance, model size, and embedding vector length. In each round $i$, the top-$K$ most relevant items $\{m^{t,i}_j\}_{j=1}^K\subset M^t$ are retrieved and provided to the agent along with the report $R^{i-1}_t$ to produce the refined report $R_t^i$. The report $R_t^N$ generated after the $N$-th round is used as the final analysis report $R_{t}$ for the news $x_t$ and to update the note as $n_{t'}$. Then it is inserted into the memory $M^t$ for future reference and pricing, updating the memory to $M^{t'}$.

The pricing network will utilize the analysis reports $\{R_{t^d_i}\}_{i=1}^{N_d}$ of all filtered news $\{x_{t^d_i}\}_{i=1}^{N_d}$ for a given day $d$, where $N_d$ represents the number of filtered news items on day $d$. Figure \ref{fig:main} provides an overview of the entire analysis process. The prompts employed by our agent are detailed in Appendix \ref{apdx:prompts}. In Section \ref{ssec:abl}, we conduct experiments on our agent design and the impact of $N$ and $K$.

\subsection{Hybrid Asset Pricing Network}
\label{ssec:hapm}

We use the embedding model $f_e$ to transform each report $R_{t^d_i}$ into an embedding $e_{t^d_i}$, where $t^d_i$ represents the timestamp of the $i$-th news on day $d$. The average embedding of the analysis reports on a given day $d$ is calculated as $e_d=\sum_{i=1}^{N_d}e_{t^d_i} / N_d$. According to \citet{NAP}, a single day's news is insufficient to fully capture the broader economic and market conditions. Therefore, we employ a sliding window of $L_W$ to derive a \textit{\textbf{smoothed daily embedding}} $s_d$ using the average embeddings of the most recent $L=min(L_W,d)$ days $\{e_{d-L+1},e_{d-L+2},...,e_d\}$ as follows:
\begin{equation*}
s_d=\sum_{i=1}^L \kappa(L,i) e_{d-L+i}
\end{equation*}
where $\kappa(L,i)$ is an exponential decay kernel defined as $\frac{\eta^{L-i}}{\sum_{j=1}^L \eta^{L-j}}$. The decay coefficient is denoted as $0<\eta<1$. We form a raw hybrid state $h_{d,a}=[s_d;v_{d,a}]$ by concatenating the smoothed daily state $s_d$ with a vector $v_{d,a}\in\mathcal{R}^{N_F}$ of $N_F$ manual-constructed financial economic factors. The asset $a$ is indexed by a permanent number (\texttt{permno}) from the Center for Research in Security Prices (CRSP) \footnote{https://www.crsp.org/} database. The hybrid state is subsequently downsampled by $h'_{d,a}=\sigma(W_Sh_{d,a})$, where $\sigma$ denotes the ReLU function and $W_S\in\mathcal{R}^{d_{model}\times (d_{emb}+N_F)}$ is a parameter matrix.

To capture the asset-specific loading to the hybrid state, especially to the asset-agnostic $s_d$, we define an asset embedding $E\in\mathcal{R}^{N_A\times d_{model}}$, which can be looked up via the permnos of the assets. Here, $N_A$ denotes the total number of assets and $d_{model}$ is the dimension of the embedding. We then concatenate the asset embedding $E_a$ with the downsampled hybrid state to form $\hat{h}_{d,a}=[h_{d,a}';\sigma(E_a)]$, the \textbf{asset-specfic hybrid state} for $a$.

The excess return of asset $a$ for the next day is predicted by $r_{d+1,a}=f_{P}(\hat{h}_{d,a})$, where $f_{P}=f_{P_{inp}}\circ f_{H_1}...\circ f_{P_{out}}$ represents a multi-layer fully connected prediction network. Specifically, $f_{P_{inp}}(\cdot)=\sigma(W_{P_{inp}}\cdot)$, with $W_{P_{inp}}\in\mathcal{R}^{2d_{model}\times d_{model}}$, and $f_{P_{out}}(\cdot)=W_{P_{out}}\cdot$, where $W_{P_{out}}\in\mathcal{R}^{d_{model}\times 1}$. Additionally, $f_{H_k}$, for $k\in[1,2,3,...]$, denotes hidden layers parameterized by $W_{H_k}\in\mathcal{R}^{d_{model}\times d_{model}}$. For simplicity, batch normalizations, residual connections, and dropout layers are not included. Figure \ref{fig:hap} illustrates the prediction network.

The hybrid asset pricing network, represented as $f_H$ and parameterized by $\theta$, comprises the embedding table $E$, the downsampling matrix $W_S$, and the prediction network $f_{P}$. We train $f_H$ using the Mean Square Error (MSE) criterion, which minimizes the average squared difference between the predicted return $r_{d+1,i}$ and the ground truth $\hat{r}_{d+1,a}$ over the training set, written as
\begin{equation*}
    \begin{split}
    \arg\min_{\theta} \frac{1}{N_AN_D}\sum_{d,a} (r_{d+1,a}-\hat{r}_{d+1,a})^2,
    \ where\ r_{d+1,a}=f_H(h_{d,a};\theta)
    \end{split}
\end{equation*}
Where $N_D$ denotes the number of days in the training set. The model is trained for $T$ episodes with a batch size of $B$. We initially pre-train this hybrid asset pricing network $f_H$ to make use of the historical factor data available before the beginning of the news dataset. During this pre-training phase, a placeholder embedding (such as the embedding for the word "Null") is utilized.

\section{Experiment}

We conduct experiments to assess the asset pricing efficacy of the proposed method. The experimental setup is detailed in Section \ref{ssec:setting}. Subsequently, we present the outcomes of the portfolio optimization experiments in Section \ref{ssec:po} and the asset pricing error in Section \ref{ssec:ape}. An extensive ablation study of our method is provided in Section \ref{ssec:abl}. Furthermore, we explore the predictive capabilities of refined news on economic indicators and stock movements in Appendix \ref{apdx:news_pred}.

\subsection{Experiment Setting}
\label{ssec:setting}

We build a dataset comprising three years of WSJ articles spanning from September 29, 2021, to September 29, 2024, based on the SocioDojo corpus \cite{sociodojo}.
This approach mitigates potential information leaks while maintaining continuity in note $n$. Besides the LLM filtering described in Section \ref{ssec:main}, we also manually excluded articles on unrelated topics like travel, lifestyle, and puzzles, based on their WSJ categories. Visualizations of our news dataset can be found in Appendix \ref{apdx:viz}. The daily asset returns are sourced from CRSP, while daily risk-free returns and market returns are obtained from Kenneth French's data library \footnote{https://mba.tuck.dartmouth.edu/pages/faculty/ken.french}.

We construct financial economic factors following \citet{jensen2023there}. In line with \citet{DLAP}, we duplicate the values from the previous time step for factors that are not updated in the current step to handle discrepancies in the update frequencies of the factors. Additionally, we imputed the missing data values using the cross-sectional median. The data split remained consistent across all our experiments: the initial 9 months of data were utilized as the training set, the following 3 months served as the validation set, and the last 1 year was reserved for testing.

We apply OpenAI \texttt{GPT-4o-0806} \cite{GPT4} as default LLM, and the latest reasoning model \texttt{O1-Preview} in addition.
Following \citet{sociodojo}, we set a low \texttt{temprature=0.2} in LLMs for all of our experiments, balancing deterministic and flexibility in analysis. 
We select five recent asset pricing baselines from highly reputed financial economics journals, validated under current empirical finance standards, as indicated by \citet{jensen2023there}, to assess our approach: \citet{NNAP} introduced a deep Neural Network for Asset Pricing (\texttt{NNAP}); \citet{IPCA} developed an Instrumental PCA (\texttt{IPCA}) to identify hidden factors and loadings; \citet{CA} proposed to use a Conditional Autoencoder (\texttt{CA}); \citet{NAP} employs LDA for the WSJ news as hidden News Factors (\texttt{NF}) similar to ours; and \citet{DLAP} utilized GAN to address Stochastic Discount Factors (\texttt{SDF-GAN}). We replicated these models using the configurations from their respective papers with their carefully chosen factor sets. For both our method and the baselines, we performed a hyper-parameter search to compare the best results. The hyper-parameter optimization setting for our method is detailed in Appendix \ref{apdx:hp}.

\subsection{Portfolio Optimization}
\label{ssec:po}

\begin{table}[]
\centering
\begin{tabular}{@{}lcccccc@{}}
\toprule
     & \multicolumn{3}{c}{SR $\uparrow$}                        & \multicolumn{3}{c}{MDD ($\%$) $\downarrow$}                   \\ \cmidrule(l){2-7} 
     & TP            & EW            & VW            & TP            & EW            & VW            \\ \midrule
NNAP   & 3.59& 2.55& 2.29& 4.89& 7.72& 8.99\\
IPCA & 3.85& 2.76& 2.48& 4.42& 6.32& 8.24\\
CA   & 3.77& 2.69& 2.35& 4.18& 7.01& 5.67\\
\textit{NF}   & \textit{3.32}& \textit{2.39}& \textit{2.12}& \textit{5.14}& \textit{7.88}& \textit{6.31}          \\
SDF-GAN  & 3.86& 2.73& 2.44& 4.77& 6.87& \textbf{5.24}\\ \bottomrule
 Ours& \underline{4.21}& \underline{3.07}& \underline{2.77}& \underline{4.00}& \underline{5.71}& 5.36\\ 

 w/ O1& \textbf{4.31}& \textbf{3.15}& \textbf{2.90}& \textbf{3.72}& \textbf{5.33}&\underline{5.27}\\ \bottomrule
\end{tabular}
\caption{Sharpe Ratio (SR) and Maximal Drawdown (MDD) for Tengency Portfolio (TP), Equal-Weighted (EW), and Value-Weighted (VW) long-short portfolio built based on the baselines and our method with the default GPT-4o-0806 or O1-Preview. We bolded the best results and underlined the second bests. }
\label{table:sr}
\end{table}

We begin by testing the Sharpe ratio for portfolios built on the predicted returns of individual assets. The Sharpe ratio (SR) \cite{sharpe1998sharpe} quantifies the risk-adjusted performance of a portfolio as $S_p=\frac{\bar{r}_p-\bar{r}_f}{\sigma(r_p)}$, where $r_f$ stands for the risk-free return, $r_p$ represents the portfolio return and $\sigma$ indicates the standard deviation. Furthermore, we evaluate the maximum drawdown, which is the largest decrease in the total value of the portfolio up to time $T$, expressed as $MDD(T)=\max_{\tau\in(0,T)}[\max_{t\in(0,\tau)} X(t)-X(\tau)]$. Here, $X(\tau)$ is the highest value, and $X(t)$ is the lowest value of the portfolio within the time interval $(0,\tau)$.

We evaluate three prevalent methods for portfolio construction. The Tangency Portfolio (TP), where the asset weights are calculated as $w_t=E_t[R^e_{t+1} {R^e_{t+1}}^{T}]^{-1} E_t[R^e_{t+1}]$, with $R^e_{t+1}$ denoting the predicted excess returns of all assets. Provides a theoretical portfolio in an ideal market without trading frictions. Next, we examine the more practical long-short decile portfolios, which involve ranking assets by their expected returns, going long on the top decile, and shorting the bottom decile. The assets in these portfolios can be either ``Equally-Weighted'' (EW) or weighted by their market capitalization, known as ``Value-Weighted'' (VW).

The experiment results are presented in Table \ref{table:sr}. Our approach achieved the highest SR across all three portfolios, with SR improvements of 9.0\%, 11.2\%, and 11.7\% respectively over the best baseline methods (SDF-GAN for TP, IPCA for EW, and VW), averaging a 10.6\% increase. Additionally, it secured the best or second-best MDD in TP and EW compared to the leading baseline IPCA with gains of 4.3\% and 9.7\%. In VW, the MDD underperforms the top baseline CA by only 2.3\%. However, substituting GPT-4o-0806 with O1-Preview \cite{jaech2024openai}, which trained with reasoning enhancements, resulted in SR improvements of 9.1\%, 14.1\%, and 12.9\% across the three portfolios, and improved MDD levels to gains of 11.0\%, 15.7\%, and -0.6\% relative to the best baselines.

\subsection{Asset Pricing Error}
\label{ssec:ape}

\begin{table}[]
\centering
\begin{tabular}{@{}lcccc@{}}
\toprule
     & avg $|\alpha|$$\downarrow$& avg $|t(\alpha)|$$\downarrow$& $\frac{\#|t(\alpha)|>1.96}{\#test\ assets}$$\downarrow$ & GRS$\downarrow$  \\ \midrule
NNAP   & 0.97& 3.02& 0.67& 6.97\\
IPCA & 0.84& 2.65& 0.60& 6.68\\
CA   & 0.89& 2.62& 0.58& 6.72\\
\textit{NF}   & \textit{0.93}& \textit{2.88}& \textit{0.63}& \textit{7.56}\\
SDF-GAN  & 0.80& 2.55& 0.59& 6.67\\ \midrule
Ours & \underline{0.72}& \underline{2.47}& \textbf{0.55}& \underline{6.48}\\ 
 w/ O1& \textbf{0.69}& \textbf{2.41}& \textbf{0.55}&\textbf{6.32}\\ \bottomrule
\end{tabular}
\caption{Asset pricing errors for anomaly portfolios with the baselines and our method with GPT-4o-0806 and O1-Preview. We bolded the best results and underlined the second bests. }
\label{table:ape}
\end{table}

We further analyze the asset pricing errors of the proposed method. Following \citet{NAP}, we chose 78 anomaly portfolios as test assets. These portfolios were constructed using 78 characteristics, including typical anomaly characteristics such as idiosyncratic volatility, accruals, short-term reversal, and others, as identified by \citet{NNAP}. We applied multiple metrics. The average absolute alpha $avg. |\alpha|$ is computed by dividing the expected value of the estimated error term $\hat{\epsilon_{t,i}}$ by the square root of the average squared returns $E[R_{t,i}]$ for all quantile-sorted portfolios. This normalization was performed to account for variations in average returns between portfolios. To measure statistical significance, we calculated the average t-value for the results and analyzed the proportion of t-values exceeding 1.96. Moreover, we conducted a Gibbons, Ross, and Shanken (GRS) test \cite{GRS} to determine if the regression intercepts, represented by $\alpha_1, \alpha_2, ..., \alpha_n$, are collectively zero. This test helps to evaluate the overall significance of the intercepts in the regression analysis.

Table \ref{table:ape} displays the results. Our method secured either the top or second-best performance among all benchmarks. It demonstrates a 10.0\% and 13.8\% reduction in average $|\alpha|$ for GPT-4o-0806 and O1-Preview respectively when compared to SDF-GAN, the leading benchmark, along with a 3.1\% and 5.5\% increase in t-value. Additionally, there is a 5.2\% reduction in the proportion of pricing results with a t-value exceeding 1.96 compared to CA for both GPT-4o-0806 and O1-Preview, as well as a 2.8\% and 5.3\% enhancement in the GRS test compared to SDF-GAN.

\subsection{Ablation Study}
\label{ssec:abl}

We conduct ablation studies to examine the influence of various components in our approach. Initially, we evaluate the performance of different modules in our agent design in Section \ref{sssec:analysis}, followed by an examination of the depth and width of the analysis, which are controlled by $N$ and $K$ respectively, in Section \ref{sssec:analysis}.

\begin{table}[!htb]
\centering
\begin{tabular}{@{}lcccc@{}}
\toprule
     & SR (EW)& MDD (EW)& avg $|\alpha|$& avg $|t(\alpha)|$\\ \midrule
\textit{NF}   &    2.39&     7.88&   \textit{0.97}&   \textit{3.02}\\
 \textit{+ Factors}& \textit{2.33}& \textit{9.12}& \textit{1.03}&\textit{3.16}\\ \bottomrule
 Naive& 2.44& 6.13& 0.95&2.77\\
+ RAG&    2.59&     5.99&   0.93&   2.71\\
+ Emb.&    2.56&     5.82&   0.92&   2.72\\ \midrule
Memory&    2.75&     7.17&   0.87&   2.69\\
 + Factors& 2.82& 6.77& 0.84&2.65\\ \midrule
Hybrid&    2.88&     6.32&   0.82&   2.49\\ 
 + Refine& 3.01& 5.89& 0.77&2.32\\
 + Notes& 2.99& 6.68& 0.76&2.53\\ \midrule
 Ours& \textbf{3.07}& \textbf{5.71}& \textbf{0.72}&\textbf{2.47}\\ \bottomrule
\end{tabular}
\caption{Ablation study of our method and comparison with NF \cite{NAP}. ``Naive'' directly produces the analysis report given news and only daily embeddings are inputted to the pricing network. ``+ RAG'' introduces the external memory and retrieves Top-K items when performing analysis for Retrieval-Augmented Generation (RAG). ``+ Emb.'' introduces the asset embeddings. ``Memory'' baseline incorporate both ``+ RAG'' and ``+ Emb.'' ``+ Factors'' introduces the daily manual factors into the pricing network in ``Memory''. ``Hybrid'' baseline pretrained the pricing network of ``Memory''. ``+ Refine'' refines the analysis report iteratively in $N$ rounds for ``Hybrid''. ``+ Notes'' introduces the macro economics and micro market notes. ``Ours'' is our method that combines ``+ Refine'' and ``+ Notes'' in ``Hybrid''.}
\label{table:abl}
\end{table}

\subsubsection{Agent Architecture Design}
\label{sssec:agent_design}

We analyze our architecture in a reverse manner, beginning with a "Naive" agent that generates the analysis report directly from the refined news without any supplementary information or iterative analysis, while the pricing network solely uses the daily embeddings as input. We then incrementally add components to develop stronger baselines until arriving at our method. The results are shown in Table \ref{table:abl}, and the baseline illustrations are provided in Appendix \ref{apdx:rag}.

Furthermore, we contrast these methods with the news-based asset pricing baseline NF \cite{NAP}, along with an NF model incorporating manual factors, akin to our full model. It is important to highlight that NF employed WSJ news over a period of 33 years, whereas we utilized only 2 years of news data.

Owing to the analytical capabilities and feature extraction proficiency of LLMs, the ``Naive'' baseline enhances the SR by 2.1\% with comparable pricing errors to NF. Incorporating external memory with RAG further boosts the SR by 6.1\% and decreases the average $|\alpha|$ by 2.1\% over ``Naive'', highlighting the significance of additional contextual information when interpreting business news. Moreover, asset embedding leads to a 4.9\% increase in SR and a 3.2\% reduction in average $|\alpha|$ by introducing asset-specific loadings.

By combining both, the ``Memory'' baseline enhances the SR of ``Naive'' by 12.7\% and decreases the average $|\alpha|$ by 8.4\% with a lower $t$ value. Incorporating the manual factors, the SR saw a slight increase of 2.5\%, while the average $|\alpha|$ decreased by 3.4\%. In comparison, the performance of NF declined after the introduction of manual factors, which is consistent with the findings of \citet{NAP}, where the inclusion of Fama-French factors reduces the SR, which may be due to suboptimal interactions between factors and news features.

After pretraining the pricing network with historical factor data, the performance of the ``Hybrid'' baseline saw a notable enhancement of 4.7\% in SR and a 5.7\% reduction in the average $|\alpha|$ when compared to the ``Memory'' baseline. This demonstrates the synergy between manual factors and LLM-generated reports, resulting in a successful non-linear interaction. The improvements from our iterative refinement and long-term notes over the ``Hybrid'' baseline are 4.5\% and 3.8\% in SR, 4.1\% and 6.1\% and 7.3\% in average $|\alpha|$, respectively with a lower $t$ value and a similar level of MDD. These enhancements collectively yield 6.6\% and 12.2\% gains in SR and average $|\alpha|$ respectively in our full method compared to a ``Hybrid'' baseline, underscoring the effectiveness of our agent architecture design.

\subsubsection{Analysis Depth and Width}
\label{sssec:analysis}

\begin{figure}
    \centering
    \includegraphics[width=0.9\linewidth]{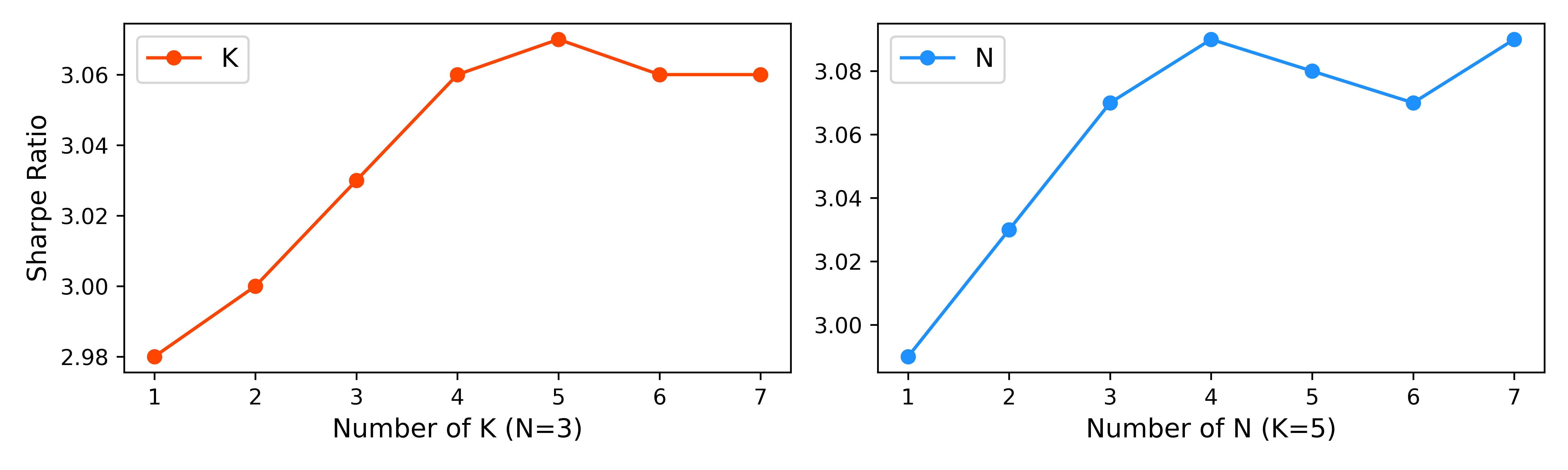}
    \caption{The Sharpe ratio of equal-weighted portfolios given different numbers of $K$ and $N$.}
    \label{fig:width-depth}
\end{figure}

We further investigate the depth of the analysis, which is controlled by the number of iterations $N$ to refine the analysis report, and the width, which is determined by $K$, the amount of relevant information to check. The results are shown in Figure \ref{fig:width-depth}. We keep one variable constant and test the other. We observe that the agent benefits from more rounds of analysis and a broader range of relevant information overall with a sharp decline in marginal gain after a certain point around $K \times N = 15$, likely due to the sufficiency of the provided information. Thus, we test an extreme case where $N=1$ and $K=15$, resulting in the SR dropping to 2.98. This indicates that iterative refinement is necessary, as items retrieved in different rounds of refinement provide diverse information as the query evolves. In contrast, a single retrieval leads to items falling into the same topic, with the value of additional items decreasing rapidly and potentially introducing noise.

\subsubsection{Foundation Model Impact}

\begin{table}[!htb]
\centering
\begin{tabular}{@{}lcccc@{}cc}
\toprule
     & SR (EW)& MDD (EW)& avg $|\alpha|$& avg $|t(\alpha)|$ & Input (\$) & Output (\$) \\ \midrule
GPT-4&    3.03&     5.57&   0.73&   2.56 & 30&60\\
 GPT-3.5& 2.89& 5.64& 0.75&2.60 & 1.0&2.0\\
 GPT-4o (Ours)& 3.07& 5.71& \underline{0.72}&\underline{2.47} & 2.5&10\\ \midrule
 O1-preview& \textbf{3.15}& 5.33& \textbf{0.69}&\textbf{2.41} & 15&60\\
O1-mini&    \underline{3.12}&     \textbf{5.12}&   \underline{0.72}&   2.56 & 1.1&4.4\\ \midrule
Llama 3.1 405B&    3.01&     \underline{5.29}&   0.73&   2.59& 3.5&3.5\\ 
 Llama 3.1 70B& 2.94& 5.91& 0.75&2.69& 0.9&0.9\\ 
 Qwen 2 72B& 2.91& 6.89& 0.77&2.67& 0.9&0.9\\ \bottomrule
\end{tabular}
\caption{Ablation study of foundation models.}
\label{table:abl_fm}
\end{table}

Table \ref{table:abl_fm} shows the analysis of how selecting different foundation models affects outcomes. We begin by contrasting widely used OpenAI models, including GPT-4o-0806 employed in our study, as well as older models such as GPT-3.5-Turbo-1106 and GPT-4-0613, to evaluate whether advanced foundational models enhance asset pricing precision.
A notable finding is a 4.6\% increase in SR when comparing GPT-3.5 to GPT-4. Although GPT-4o shows minimal disparity with GPT-4, it is significantly more cost-effective. Our findings indicate that our method can benefit from the evolution of LLMs, possibly yielding better outcomes with newer models over time.
Importantly, the outdated models in this experiment have a knowledge cut-off preceding our dataset, thereby eliminating any information leakage risk, yet they still surpass the baselines; specifically, GPT-3.5 and GPT-4 improve SR by 5.9\% and 11.0\% compared to the leading SDF-GAN and improve avg. $|\alpha|$ by 6.3\% and 8.8\% respectively. 

Additionally, recent progress in reasoning models, such as O1-preview and O1-mini, demonstrates a 2.6\% and 1.6\% enhancement in SR compared to standard chat models like GPT-4o. Experiments also include open-source models via together.ai \footnote{\url{https://www.together.ai/}}, with notable contenders being Llama 3.1 and Qwen 2 across various sizes. Of these, the formidable Llama 3.1 405B rivals GPT-4 at a reduced token cost, and smaller variants around 70B surpass the older GPT-3.5, again offering a cost advantage. These findings underline the promise of high-efficiency local deployment using our method.

\section{Discussion}
\label{sec:disc}

Our proposed approach presents a promising method to fuse qualitative discretionary investment with quantitative factor-based strategies through the use of LLM agents in asset pricing. Nonetheless, there is still much to investigate regarding additional capabilities of LLM agents that could further enhance asset pricing power. Firstly, internet access and a broader range of information sources, including those available in SocioDojo, may enable the agent to generate more in-depth analyses, as discretionary investment relies on information beyond just news or domain knowledge. Secondly, employing specialized financial LLMs like FinGPT \cite{yang2023fingpt} could further improve the agent's financial analytical capabilities. Finally, it is crucial to consider multimodal information, such as diagrams and figures, which are frequently presented in financial documents.

\section{Conclusion}

In this research, we introduced an LLM agent asset pricing model, a model that combines qualitative analysis from the LLM agent with quantitative factors in asset pricing. Our method surpassed established asset pricing methods in multiple evaluations, including portfolio optimization and asset pricing error. Additionally, we performed an in-depth analysis of each component in our agent design. 
We posit that our research enhances the understanding of how discretionary investment strategies intersect with quantitative factor-driven models, providing evidence for the effective application of LLMs in asset pricing, and contributing to societal improvements in economic efficiency.



\section*{Ethics Statement}
We do not identify any ethical concerns in our approach. Our study does not involve any human participation. Furthermore, the application area of our method is not directly related to humans, reducing the risk of abuse or misuse. In fact, considering a wider range of information, our method has the potential to enhance market efficiency, resulting in economic benefits for society.

\section*{Reproducibility Statement}
Comprehensive descriptions of our experimental configuration are presented in Section \ref{ssec:setting}, while details pertaining to hyperparameter optimization are provided in Appendix \ref{apdx:hp}. The ablation study methodology is elaborated upon in Appendix \ref{apdx:rag}. We document all utilized resources, such as the CRSP database and the together.ai platform. Subsequent to publication, our code and data for replicating the study outcomes will be made accessible.

\bibliography{iclr2025_conference}
\bibliographystyle{iclr2025_conference}

\appendix

\section{Hyperparam Search}
\label{apdx:hp}

\begin{table}[H]
\centering
\begin{tabular}{@{}ll@{}}
\toprule
Parameter & Distribution \\ \midrule
          Learning rate&              \{1e-3,1e-4,5e-4,5e-3\}\\
          $d_{model}$&              \{128,256,512,768,1024\}\\
          $d_{emb}$&              \{128,256,512,768,1024\}\\
          Epochs&              \{50,100,150,200\}\\
          Hidden Layers&              \{0,1,2,3,4,5\}\\
          Dropout rate&              $U(0,0.3)$\\
          Batch size&              $U_{log}(32,1024,8)$\\
          $\eta$&              $U(0.9,1)$\\
 $L_W$&\{1,7,15,30,45,60,90,180\}\\
          $N$&				\{1,2,3,4,5\}\\
          $K$&				\{1,2,3,4,5\}\\
          \bottomrule
\end{tabular}
\caption{Distributions for the key hyperparameters in the hyperparameter search.}
\label{tab:hp}
\end{table}

For our approach, we conduct hyperparameter searches using Weights \& Biases Sweep \citep{wandb}. Table \ref{tab:hp} shows the distribution of empirically significant parameters used for our hyperparameter search. Here, $U(a,b)$ signifies a uniform distribution between $a$ and $b$, while $U_{log}(a,b,r)$ indicates a logarithmic uniform distribution with base $r$ between $a$ and $b$. The evaluation criteria of our method are based on the Sharpe ratio of an equal-weight long-short portfolio.

We conducted our experiments on our clusters, the major workload has the following configuration:
\begin{itemize}
    \item 2 $\times$ Intel Xeon Silver 4410Y Processor with 12-Core 2.0GHz 30 MB Cache
    \item 512GB 4800MHz DDR5 RAM
    \item 2 $\times$ NVIDIA L40 Ada GPUs (no NVLink)
\end{itemize}
We employed PyTorch Lightning \citep{Falcon_PyTorch_Lightning_2019} for parallel training.

\section{Illustration of the Ablation Baselines}
\label{apdx:rag}

\begin{figure*}[!htb]
	\centering
	\includegraphics[width=\linewidth]{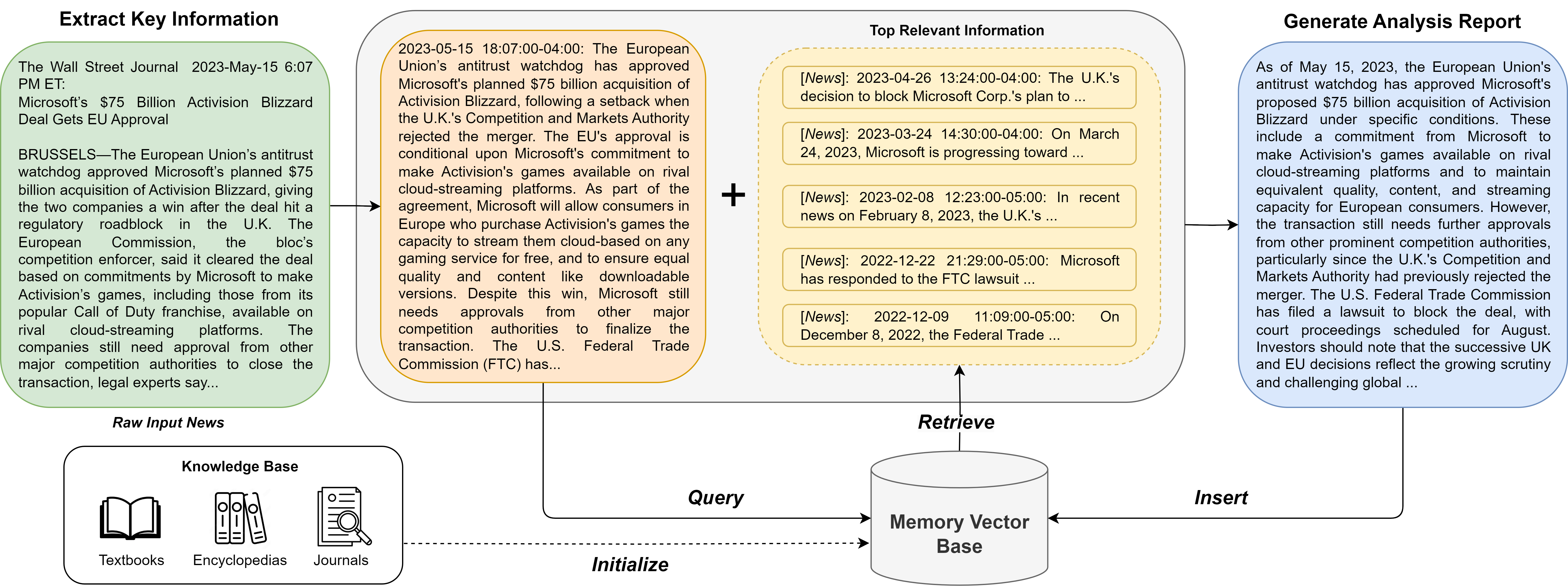}
	\caption{Example of the Hybrid agent baseline that analyzes raw news without iterative refinement of analysis report as well as the macroeconomic and market trend notes.}	\label{fig:rag_agent}
\end{figure*}

We progressively developed three baselines, starting with a naive agent, followed by a memory agent enhanced with an external vector base, and culminating in a hybrid agent that incorporated manual factors as discussed in Section \ref{ssec:abl}. Figure \ref{fig:rag_agent} illustrates an example of how a hybrid agent generates an analysis report from raw news input without iterative refinement. The analysis report is generated directly using the top-5 relevant items from the memory with the following prompt:

\begin{beautifulbox}
{\fontfamily{\myfont}\selectfont \footnotesize
You are a helpful assistant designed to analyze the business news to assist portfolio management.
Now, read this latest news and summarize it in one single paragraph, preserving data, datetime of the events, and key information, and include new insights for investment using the recommended relevant information:

{\color{blue} \{news\}}
}
\end{beautifulbox}

The architecture of the agent for the memory baseline mirrors that of the hybrid baseline. In the naive baseline, external memory is omitted, and the analysis report is produced directly from the refined news. The pricing network for the hybrid agent is identical to our method depicted in Figure \ref{fig:hap}, while the memory baseline omits the middle branch of manual factors. The naive baseline additionally removes the asset embedding branch.

\section{Analysis on Dataset}
\label{apdx:viz}

In this section, we analyze the dataset with a focus on the first two years, a highly complicated period as illustrated in Section \ref{sec:method} to provide better insights on the news serves as market signals. 

\begin{figure}[!htb]        
\centering
        \includegraphics[width=0.5\linewidth]{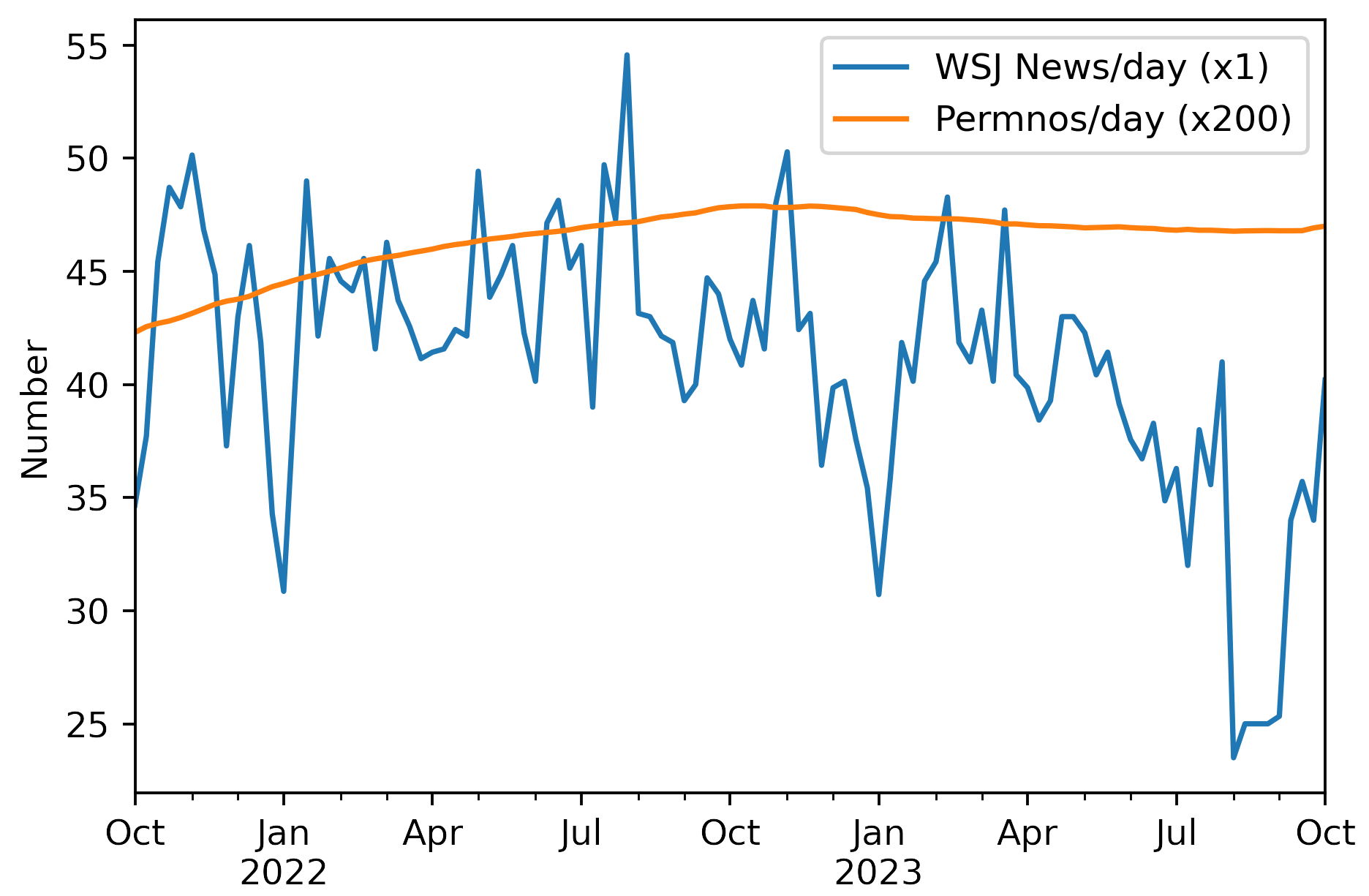}
        \caption{The number of filtered WSJ articles and active assets per day.}
    \label{fig:viz_data}
\end{figure}

\begin{figure*}[!htb]
    \centering
    \includegraphics[width=\linewidth]{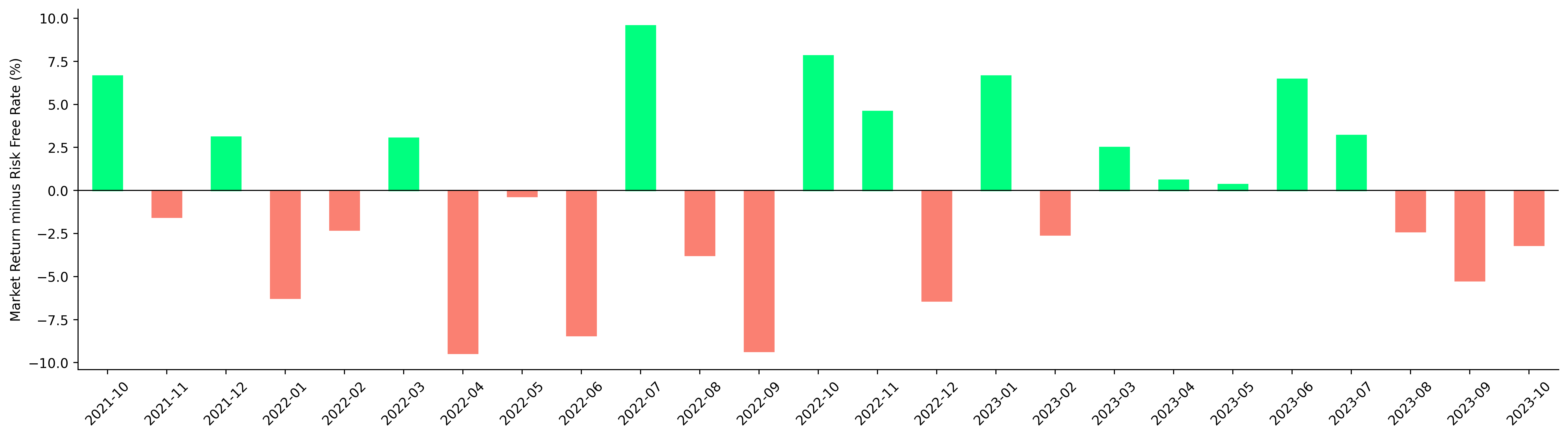}
    \includegraphics[width=\linewidth]{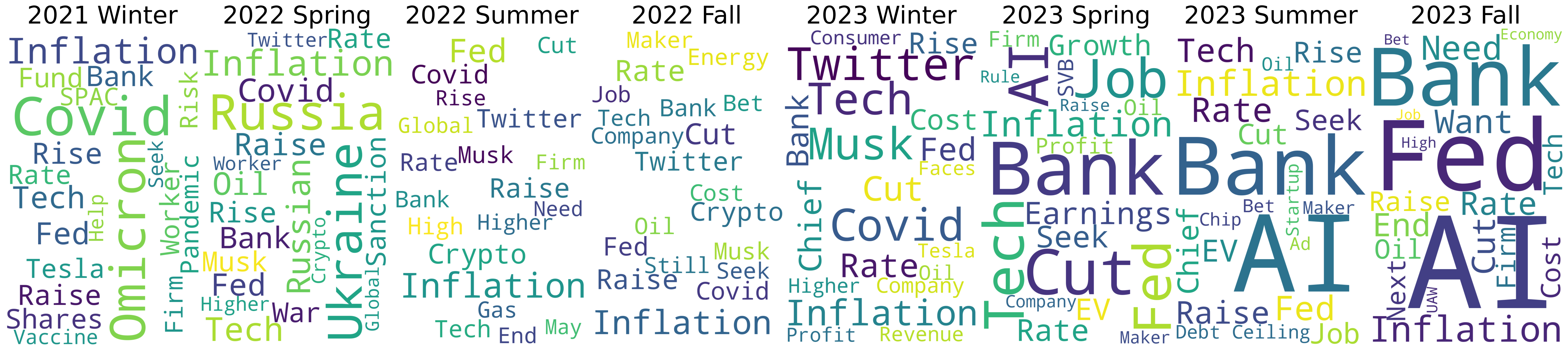}
    \caption{The word cloud of topics of the WSJ business news over the first two years in our dataset (Bottom) compared to the corresponding risk-adjusted market return (Top).}
    \label{fig:viz_topic}
\end{figure*}

\paragraph{News and Topics over time.}
Figure \ref{fig:viz_data} illustrates the variations in the number of articles and assets over time for the first two years. We analyze the primary topics discussed in the news articles within our dataset across different periods. The topics were determined based on the titles of the news articles for each season. Common words such as "US," "Stock," and "Market" were excluded as they did not effectively represent the event's topic. The resulting word cloud is shown in Figure \ref{fig:viz_topic}. It is clear that the economy is mainly influenced by various long-term events. It begins with a gradual decline in the emphasis on COVID. Then, the focus shifted towards managing inflation and the decisions made by the FED. The banking crisis at the start of 2023 soon became the new central point, followed by the acknowledgment of AI as a key driver for the economy, mainly due to the success of LLMs. This indicates that these event trends have the potential to serve as strong predictors of economic indicators and the market. This is also reflected in Appendix \ref{apdx:news_pred}, where we evaluated that news articles have significant predictive power for economic indicators and market trends.

\paragraph{Ticker Frequencies.}
We then use GPT to analyze the relevant tickers for each news item in our dataset with the following prompt:

\begin{beautifulbox}
{\fontfamily{\myfont}\selectfont \footnotesize
You are a helpful assistant designed to analyze the business news.
You need to extract the stock tickers of the companies most closely related to the news. If there is no relevant ticker, return an empty list. You should never make up a ticker that does not exist.
Now, analyze the following news: {\color{blue} \{input\}}
}
\end{beautifulbox}

\begin{figure}
    \centering
    \includegraphics[width=0.45\linewidth]{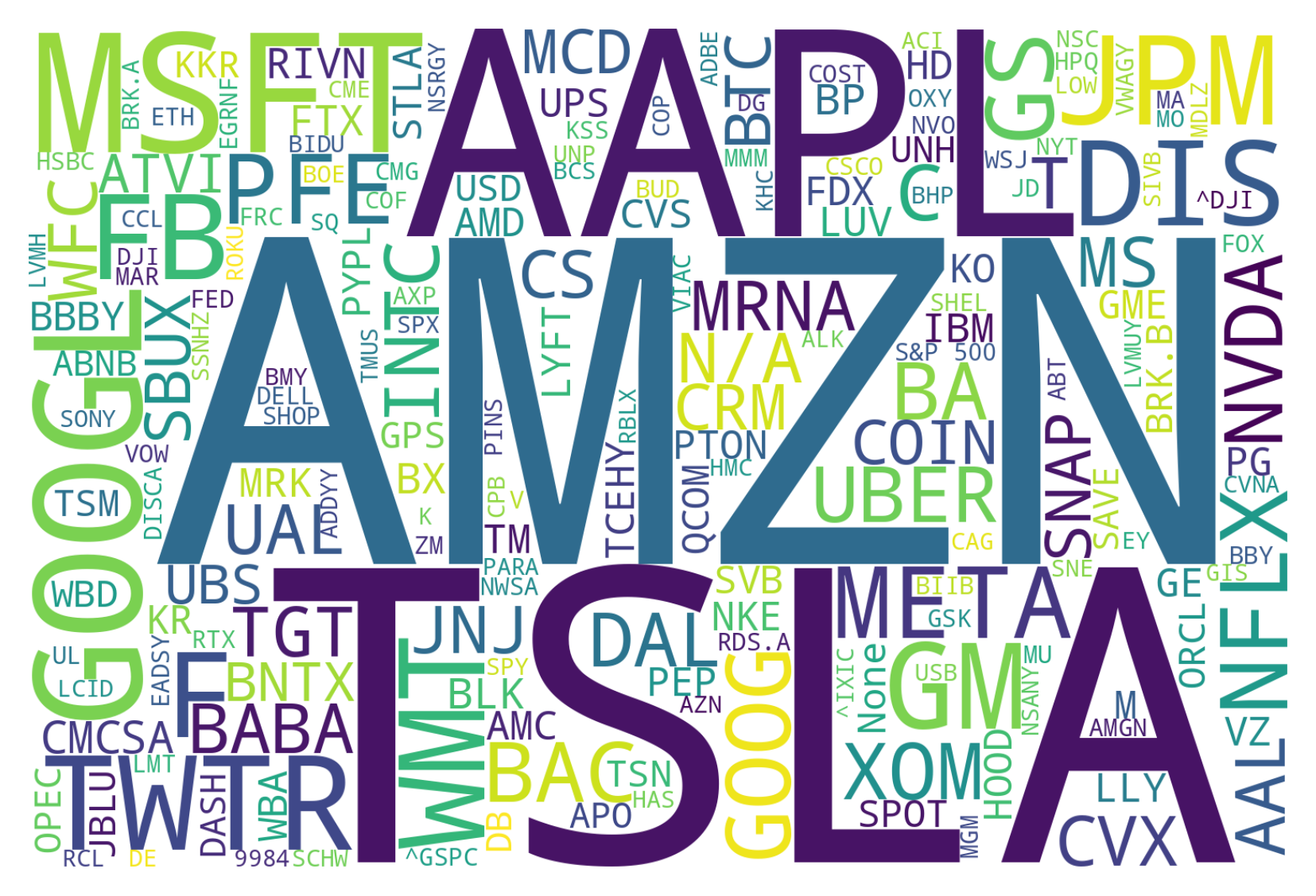}
    \caption{The most frequently mentioned stock tickers in the first two years.}
    \label{fig:ticker_viz}
\end{figure}

The stock tickers linked to the news in our dataset are displayed in Figure \ref{fig:ticker_viz}. Over the two-year span, technology stocks have evidently been the market's primary focus, aligning with our impression and the actual robust performance of these stocks over the period.

\section{Additional Results}
\label{apdx:news_pred}

We test the capacity of our method to explain and predict the excess returns of assets at different levels of performance in the highly complicated market environment in Section \ref{ssec:decile}.
To explore the predictive capability of business news on financial and economic dynamics, we conduct an experiment using refined news features to forecast the economic indicators in Appendix \ref{ssec:nep} and market movements in Appendix \ref{ssec:ticker}. We embed the refined news directly and use the daily averaged embeddings of the refined news as predictors in our experiments.

\subsection{Decile Portfolios}
\label{ssec:decile}

        \begin{figure}[!htb] 
            \centering
            \includegraphics[width=0.5\linewidth]{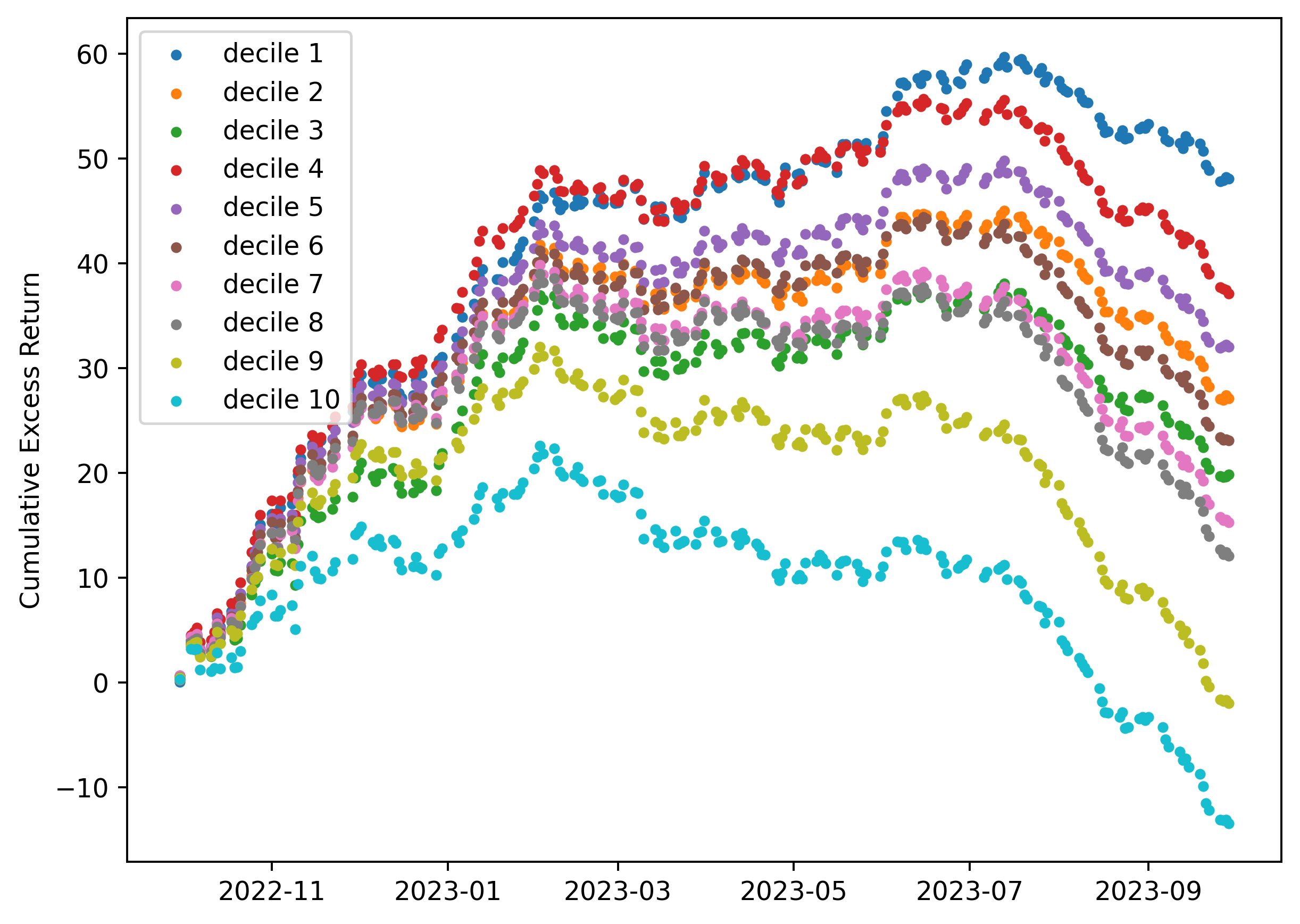}
            \caption{Cumulative excess return for decile portfolios.}
            \label{fig:cumu_er}
        \end{figure}

We assess the proposed method in explaining excess returns of assets from different performance levels by applying it to the pricing of decile portfolios. We highlight the first two-year span which shows a highly complicated market environment involving factors like COVID, inflation, the war in Ukraine, and the beginning of the AI explosion as demonstrated in Appendix \ref{apdx:viz}.
This process includes sorting the assets according to their predicted returns and then forming portfolios for each decile. Figure \ref{fig:cumu_er} shows the cumulative excess return over time. The figure clearly demonstrates that each decile creates a distinct ranking of returns in the right position, suggesting that the proposed approach effectively predicts returns at various levels.

\subsection{News as Economic Indicators}
\label{ssec:nep}

\begin{figure*}[!htb]
    \centering
    \includegraphics[width=0.9\linewidth]{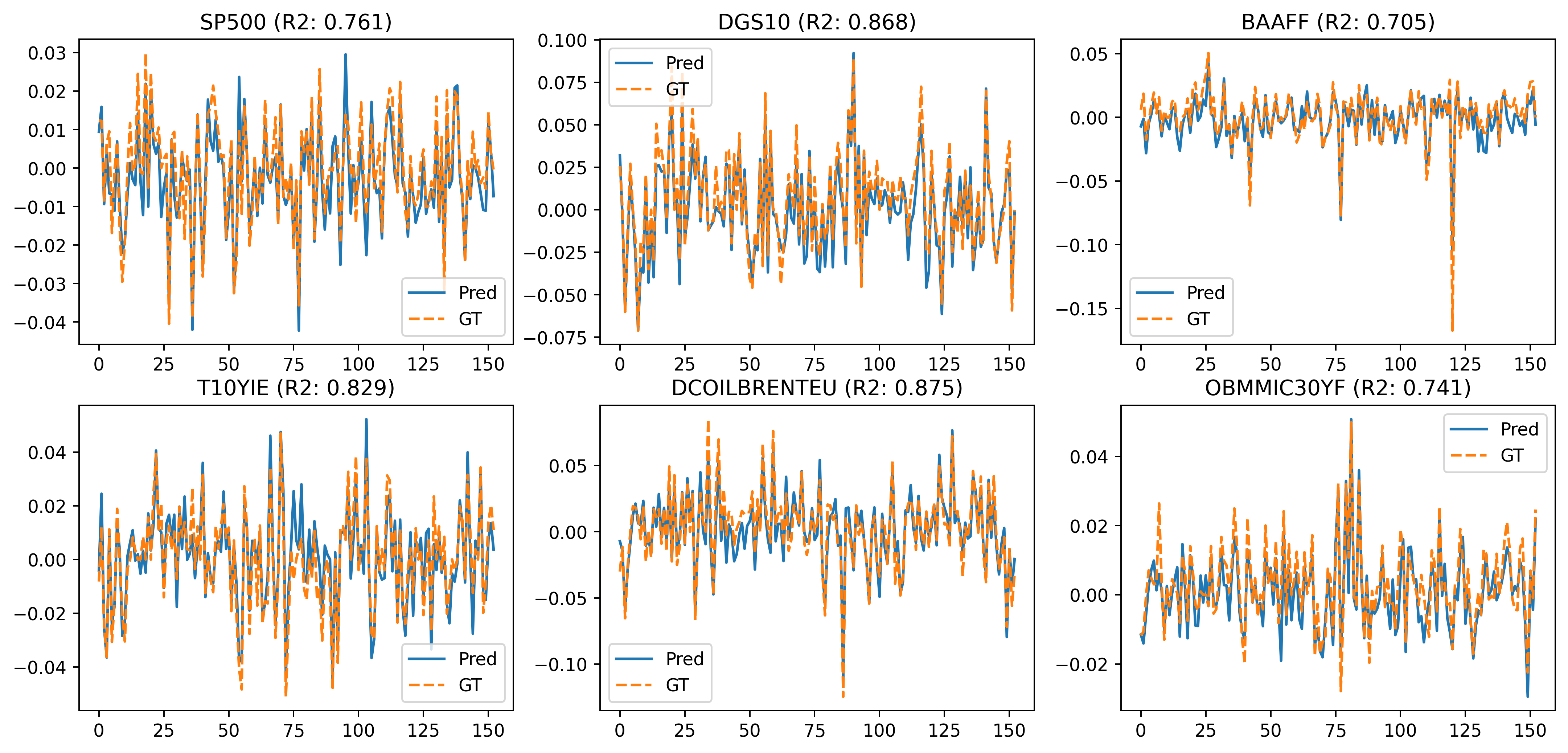}
    \caption{Use news features as the predictor to predict daily percentage change of the economic indicators.}
    \label{fig:pred_econ}
\end{figure*}

We assess the capability of news-based features to forecast the daily percentage changes in typical and most popular macroeconomic indicators in different topics sourced from the FRED database\footnote{https://fred.stlouisfed.org/}. These indicators encompass the stock market (SP500), the market yield on U.S. Treasury Securities at a 10-year constant maturity (DGS10), Moody's seasoned Baa corporate bond minus the federal funds rate (BAAFF), the 10-year breakeven inflation rate (T10YIE), Brent crude oil prices (DCOILBRENTEU), and the 30-year fixed-rate conforming mortgage index (OBMMIC30YF). The findings are illustrated in Figure \ref{fig:pred_econ}. The forecasted results exhibit a high degree of accuracy, as evidenced by the high R2 score. This implies that news provides valuable insights for predicting macroeconomic indicators.

\subsection{News as Stock Price Predictor}
\label{ssec:ticker}

\begin{figure*}[!htb]
    \centering
    \includegraphics[width=\linewidth]{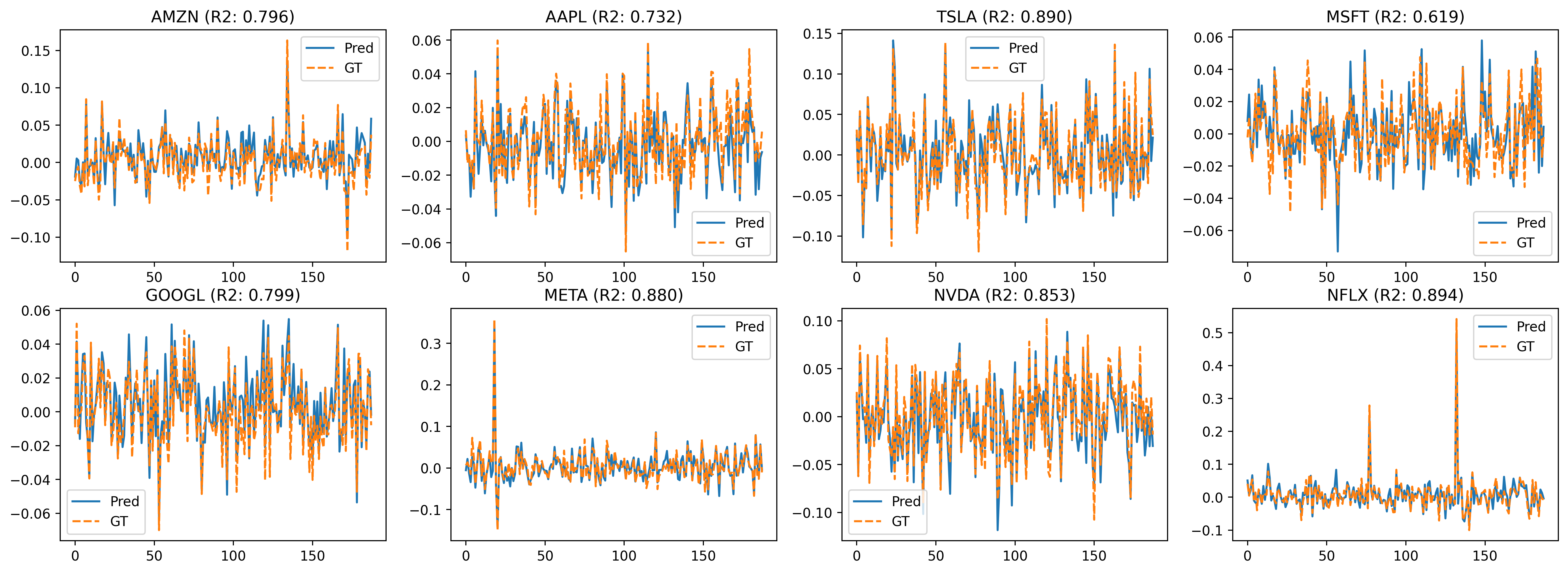}
    \caption{Predict the price movement of stocks in focus using augmented news features as predictors.}
    \label{fig:ticker_pred}
\end{figure*}

We further investigate the predictive power of news-based features to the price movements of individual stocks. We chose 8 typical stocks that have been frequently mentioned in the news from our analysis in Appendix \ref{fig:ticker_viz} and used refined news features as predictors to estimate their daily percentage price changes. The results are displayed in Figure \ref{fig:ticker_pred}, with the corresponding R2 scores given in brackets. We note high accuracy and R2 scores for all selected stocks, suggesting that news can significantly help in forecasting stock prices. However, it is important to recognize that due to non-stationarity and the risk of overfitting, stock price prediction cannot be directly applicable as asset pricing \cite{FML}, but it provides insights into the value of the news in the pricing of individual stock.

\section{Prompts}
\label{apdx:prompts}

In this section, we will present the prompts utilized by the agent, covering the refinement of the raw news input, the iterative refinement of the analysis report, the initial macroeconomic note, and the updating of notes.

\subsection{News refinement}

This refinement of the raw news input discussed in Section \ref{ssec:main} is achieved through the following prompt:

\begin{beautifulbox}
{\fontfamily{\myfont}\selectfont \footnotesize
You are a helpful assistant designed to analyze business news. You need to use brief language to describe key information and preserve key data in the news. Now, analyze the following news: {\color{blue} \{input\}}
}
\end{beautifulbox}

\subsection{Iterative analysis}

In the first iteration, the analysis begins with the following prompt:

\begin{beautifulbox}
{\fontfamily{\myfont}\selectfont \footnotesize
You are a helpful assistant designed to analyze the business news to assist portfolio management. 
You will help me analyze this latest news from The Wall Street Journal and provide an analysis report, then I will search the relevant news or articles from the knowledge base based on your analysis report to help you refine iteratively in multiple rounds. 
Let's start with this latest news, provide your analysis report, and I will help you refine it with the relevant information later, if you think this news is completely not helpful for investment now or future, call the skip function to skip it, do not skip it if it may contain helpful information to future investment: 

{\color{blue}\{inputs\}}

Here is a summary of the macroeconomics by today and the investment notes:

{\color{blue}\{macro\}}
}
\end{beautifulbox}

After the first iteration, the agent will be prompted as follows to continue the analysis:

\begin{beautifulbox}
{\fontfamily{\myfont}\selectfont \footnotesize
Based on your current analysis report, I found potentially relevant news and excerpts from the knowledge base, please refine your analysis report with this information:

{\color{blue}\{inputs\}}
}
\end{beautifulbox}

In the last iteration, the agent will end the analysis with this prompt:

\begin{beautifulbox}
{\fontfamily{\myfont}\selectfont \footnotesize
Based on your current analysis report, I found potentially relevant news and excerpts from the knowledge base, now finish your analysis report with them:

{\color{blue}\{inputs\}}
}
\end{beautifulbox}

\subsection{Update of note}

The macroeconomic and market trend note is updated with the new analysis report by the following prompt:

\begin{beautifulbox}
{\fontfamily{\myfont}\selectfont \footnotesize

Here is the current summary of the macroeconomic landscape and investment notes as of {\color{blue}\{date\}}:

{\color{blue}\{macro\}}

Now, given the latest news and the analysis report, update the macroeconomic summary with the new insights and impacts from the news. Include any relevant information that could influence the global economic outlook, such as geopolitical events, policy changes, or economic indicators. 
You should also take note of any important notes about investment trends and chances. Here are the latest news and the analysis report:

{\color{blue}\{news\}}

Now, update the macroeconomic summary with the new insights and impacts from the news as well as the investment notes.

}
\end{beautifulbox}

\subsection{Initial Macroeconomic Summary}

We use the LLM to summary the macroeconomic status before the beginning time of the dataset, the summarization generated by GPT-3.5-Turbo-1106 is as follows:

\begin{beautifulbox}
{\fontfamily{\myfont}\selectfont \footnotesize
By September 2021, the global macroeconomic landscape was heavily influenced by the ongoing impacts of the COVID-19 pandemic. Many countries were in various stages of recovery, grappling with challenges such as disrupted supply chains, inflationary pressures, and shifts in employment patterns. Key points include:
1. **Economic Recovery**: Different regions experienced uneven recovery, with some economies bouncing back faster due to successful vaccination campaigns and substantial fiscal stimuli. For instance, the U.S. and China showed signs of robust economic rebound, whereas many European countries were still struggling with economic output below pre-pandemic levels.
2. **Inflation Concerns**: Rising inflation became a significant concern in many countries, partly due to supply chain disruptions and increased demand as economies reopened. This led to higher prices for commodities, goods, and services.
3. **Monetary Policy**: Central banks, including the U.S. Federal Reserve and the European Central Bank, maintained accommodative monetary policies, with low interest rates to support economic growth. However, there was growing discourse about when and how to start tapering these measures.
4. **Employment Fluctuations**: While some sectors and countries saw a rapid recovery in employment levels, others faced ongoing job losses, highlighting the pandemic's uneven impact across different industries.
5. **Supply Chain Disruptions**: Global supply chains were strained, impacting everything from consumer electronics to automobile manufacturing, leading to shortages and delays.
6. **Shifts in Consumer Behavior**: The pandemic accelerated trends like online shopping and remote working, reshaping economic activities and consumer behaviors in lasting ways.
Overall, the state of global macroeconomics by September 2021 was defined by recovery efforts amidst ongoing challenges, with significant variability between different countries and regions.
}
\end{beautifulbox}

\end{document}

%% file: math_commands.tex
\usepackage{amsmath,amsfonts,bm}

\def\eqref#1{equation~\ref{#1}}

\def\1{\bm{1}}

\DeclareMathAlphabet{\mathsfit}{\encodingdefault}{\sfdefault}{m}{sl}
\SetMathAlphabet{\mathsfit}{bold}{\encodingdefault}{\sfdefault}{bx}{n}

